\documentclass[conference]{IEEEtran}
\usepackage{cite}
\usepackage{amsmath,amssymb,amsfonts}
\usepackage{algorithmicx}
\usepackage{graphicx}
\usepackage{textcomp}
\usepackage{xcolor}
\def\BibTeX{{\rm B\kern-.05em{\sc i\kern-.025em b}\kern-.08em
    T\kern-.1667em\lower.7ex\hbox{E}\kern-.125emX}}

\usepackage[draft=false]{defs-common}
\usepackage{defs-custom}
\usepackage{multirow}
\usepackage{hyperref}
\usepackage[noend]{algpseudocode}

\title{On Linking Level Segments}

\author{
{Colan F. Biemer}\\
\IEEEauthorblockA{Northeastern University \\
biemer.c@husky.neu.edu}
\and
\IEEEauthorblockN{Seth Cooper}
\IEEEauthorblockA{Northeastern University \\
se.cooper@northeastern.edu}
}

\begin{document}

\maketitle

\begin{abstract}
An increasingly common area of study in procedural content generation is the creation of level segments: short pieces that can be used to form larger levels. Previous work has used concatenation to form these larger levels. However, even if the segments themselves are completable and well-formed, concatenation can fail to produce levels that are completable and can cause broken in-game structures (e.g. malformed pipes in \textit{Mario}). We show this with three tile-based games: a side-scrolling platformer, a vertical platformer, and a top-down roguelike. To address this, we present a Markov chain and a tree search algorithm that finds a link between two level segments, which uses filters to ensure completability and unbroken in-game structures in the linked segments. We further show that these links work well for multi-segment levels. We find that this method reliably finds links between segments and is customizable to meet a designer’s needs.
\end{abstract}


\newcommand{\XFIGUREplaceholder}{
\begin{figure}[t]
\centering
\includegraphics[width=0.975\columnwidth]{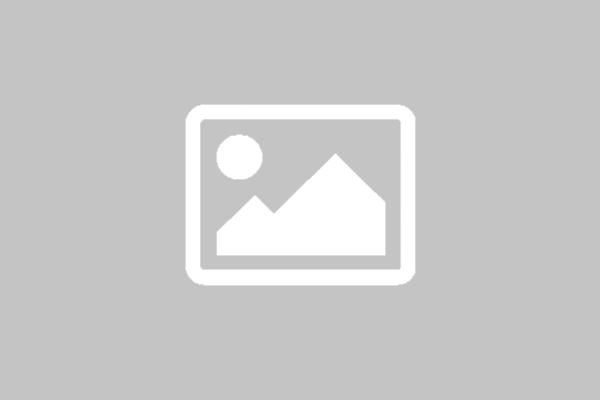}
\caption{\label{XFIGUREplaceholder} Placeholder figure.}
\end{figure}
}

\newcommand{\XFIGURELinkLengths}{
\begin{figure}[t]
\centering
\includegraphics[width=0.975\columnwidth]{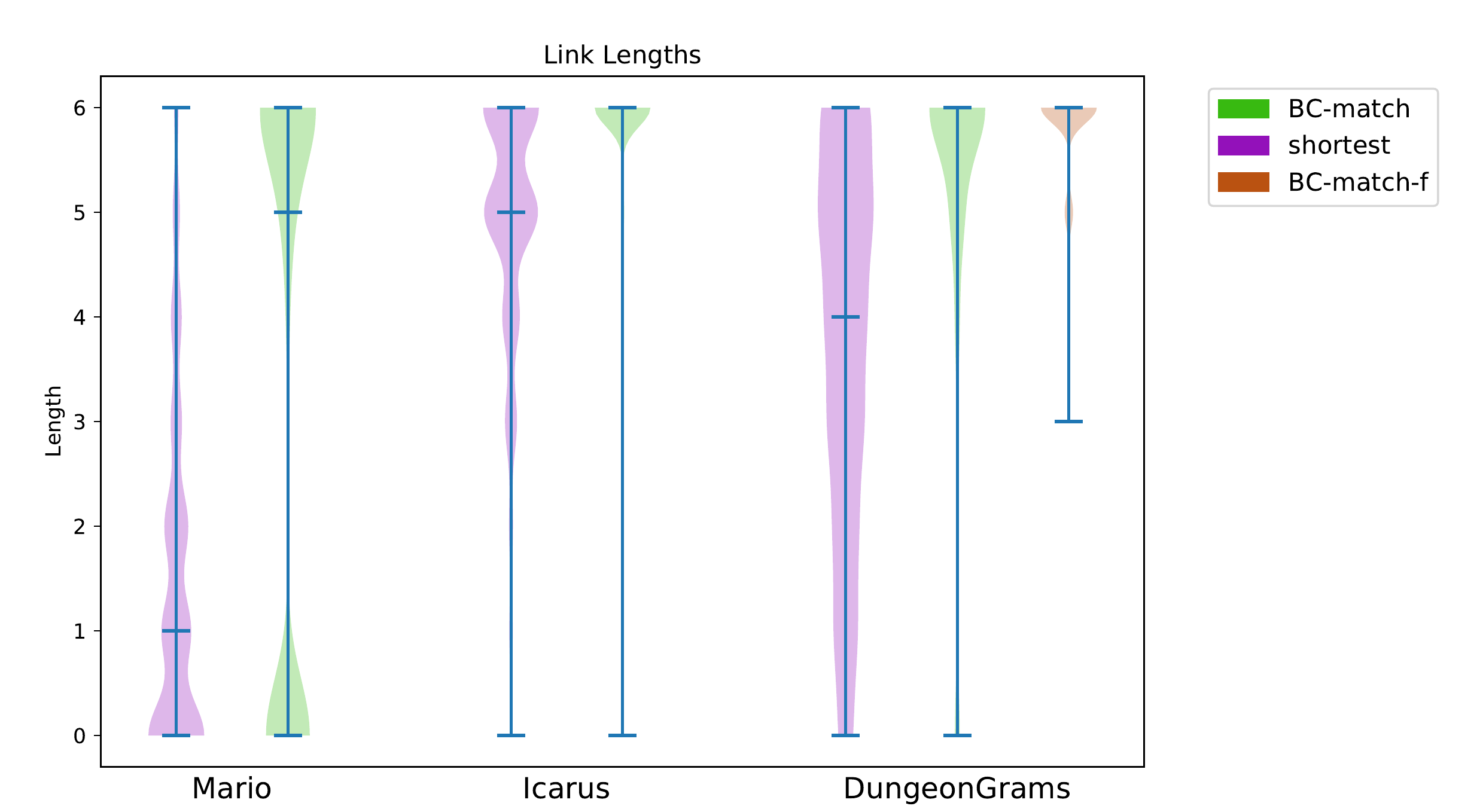}
\caption{\label{XFIGURELinkLengths} Linker lengths created for each game for the non-null linkers.}
\end{figure}
}

\newcommand{\XFIGUREIcarusLinkingColumns}{
\begin{figure}[t]
\centering

\begin{tabular}{c}
     \includegraphics[width=1.5in]{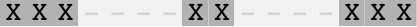} \\
     \includegraphics[width=1.5in]{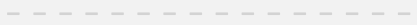} \\
     \includegraphics[width=1.5in]{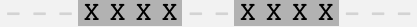} 
\end{tabular}

\caption{\label{XFIGUREIcarusLinkingColumns} The three linking slices used for for \textit{Kid Icarus}.}
\end{figure}
}

\newcommand{\XfigureBC}{
\begin{figure*}
\centering
\includegraphics[width=0.8\textwidth]{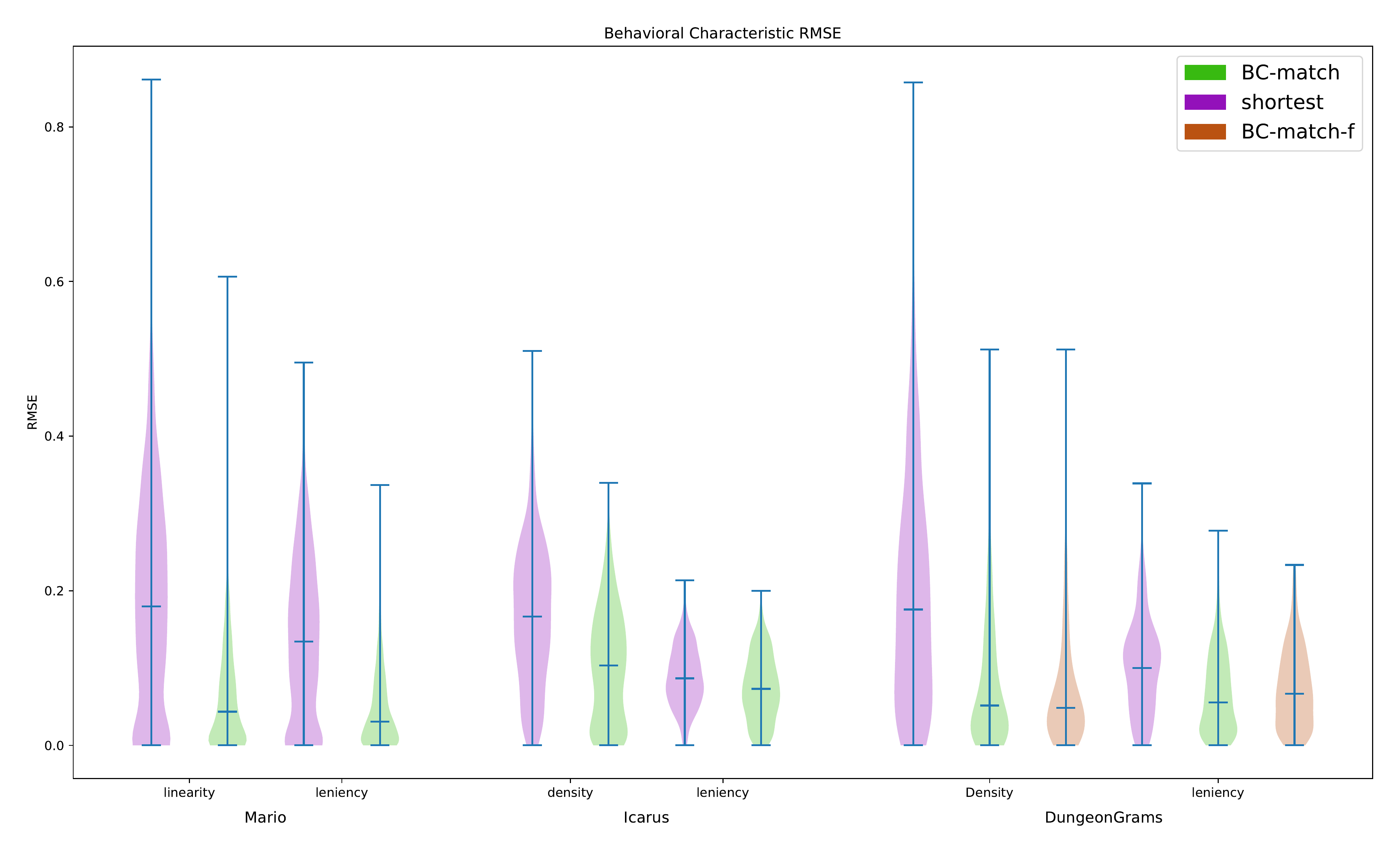}
\caption{\label{XFIGUREBC} Violin plot of the RMSE for each individual metric for each game.}
\end{figure*}
}

\newcommand{\XfigureBCSingular}{
\begin{figure}
\centering
\includegraphics[width=0.975\columnwidth]{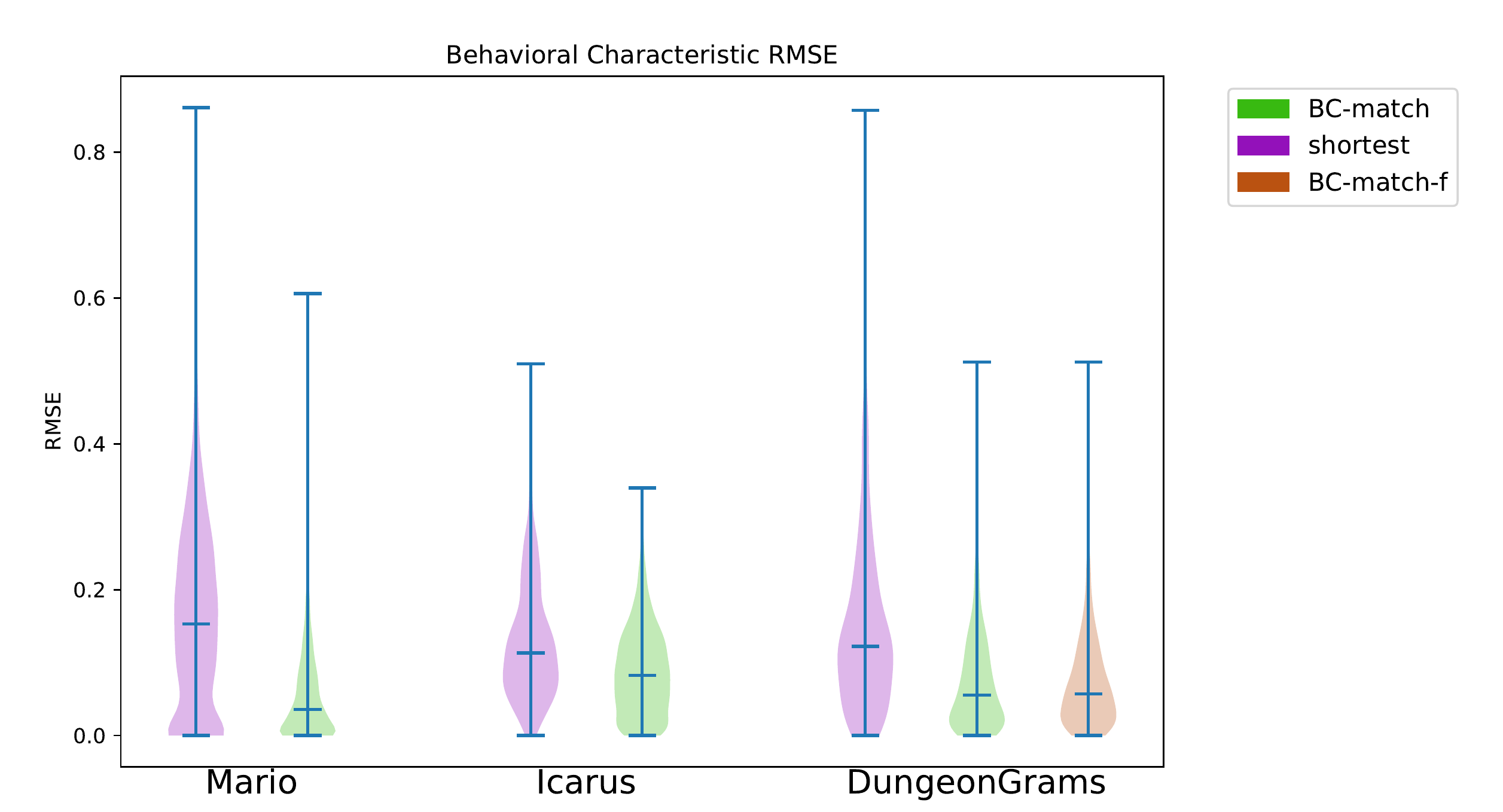}
\caption{\label{XfigureBCSingular} Violin plot of the RMSE for for each game for the non-null linkers.}
\end{figure}
}

\newcommand{\XFigureGoodLink}{
\begin{figure}
    \centering
    \includegraphics[width=0.8\columnwidth]{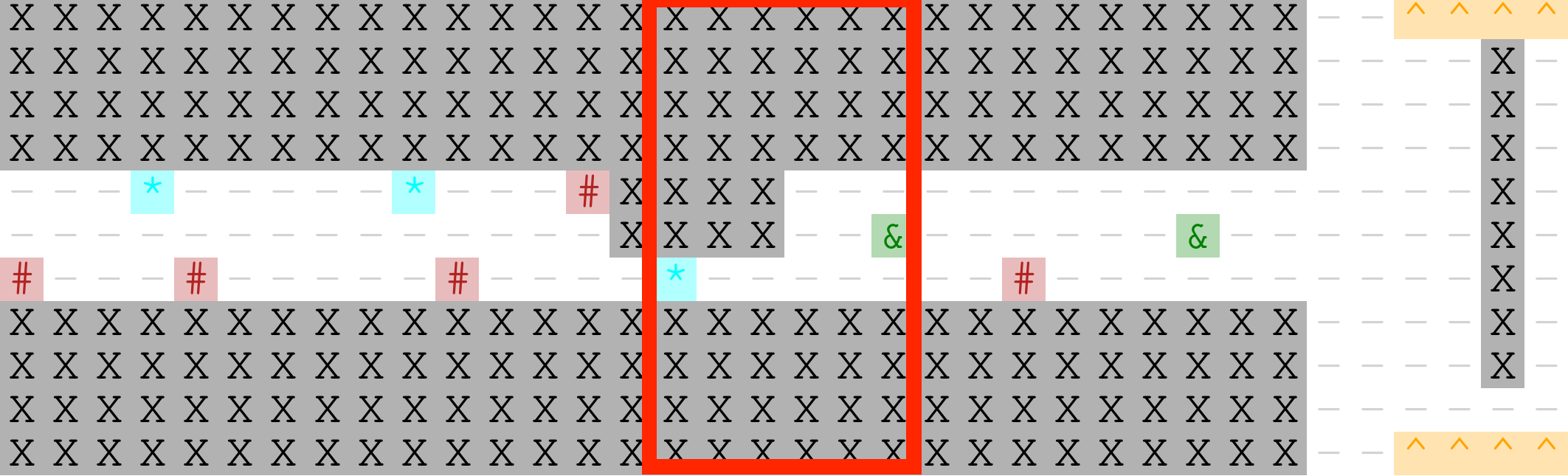}
    \caption{\label{XFigureGoodLink} Two examples of a linker generated in \textit{DungeonGrams}. The linker---generated by all three n-gram linkers---in the red box includes food (green ampersand) that can be reached, which allows the agent to complete the level. A null linker fails to produce a completable level for the two surrounding segments due to stamina.}
\end{figure}
}

\newcommand{\XfigureMalformedPipe}{
\begin{figure}
    \centering
    \begin{tabular}{cc}
    \includegraphics[height=1.1in]{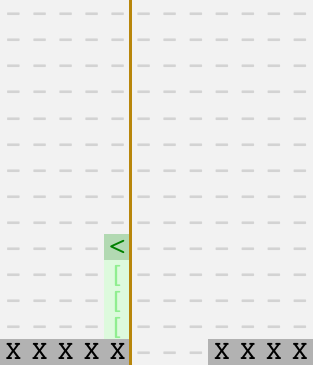}
    &
    \includegraphics[height=1.1in]{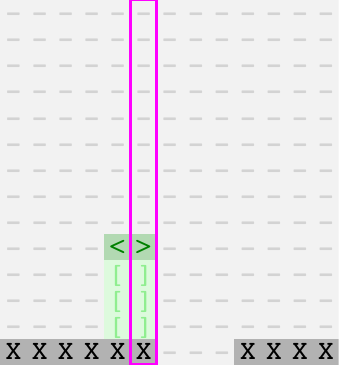}
    \\
    (a) & (b)
    \end{tabular}
    
    \caption{\label{XfigureMalformedPipe} (a) The tan line marks where the starting segment ends and the end segment begins. There is an incomplete pipe with concatenation. (b) The magenta box shows the output of the forward chain which completes the pipe.}
\end{figure}
}

\newcommand{\XFigureMarioLinkers}{
\begin{figure}
    \centering
    \small
    \begin{tabular}{cl}
      \rotatebox{90}{Null}  &  \includegraphics[height=0.7in]{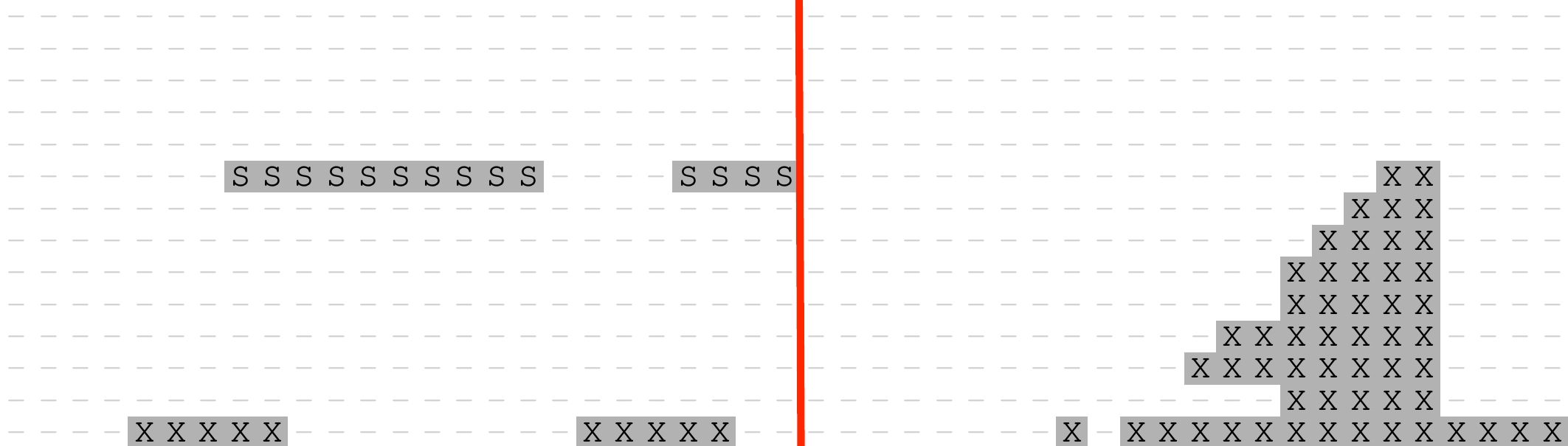}
      \\
      \rotatebox{90}{Shortest}  &  \includegraphics[height=0.7in]{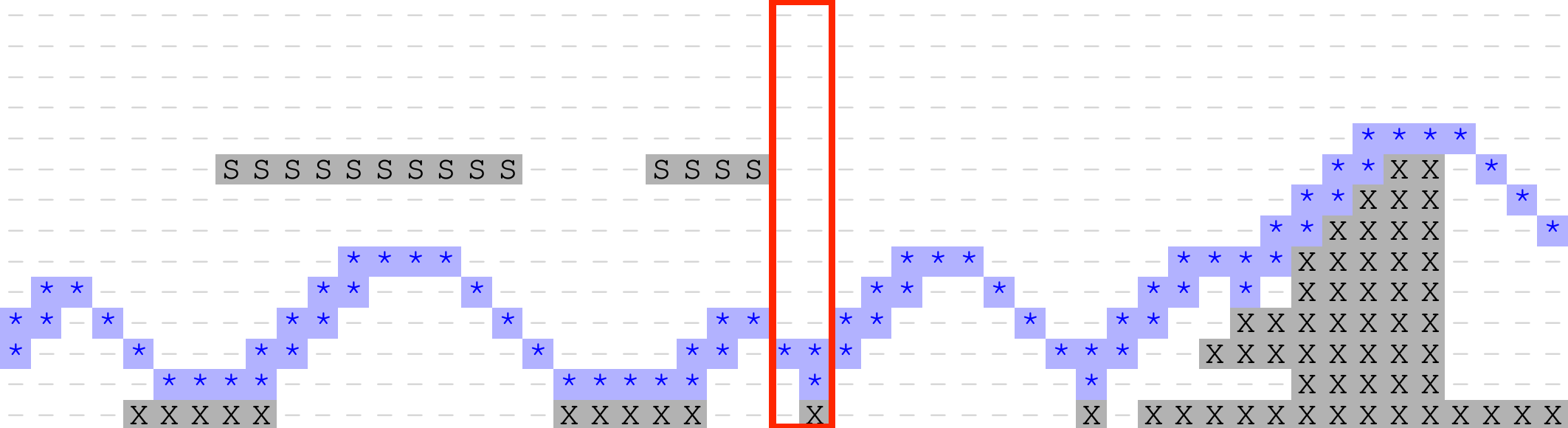}
      \\
      \rotatebox{90}{BC-match}  &  \includegraphics[height=0.7in]{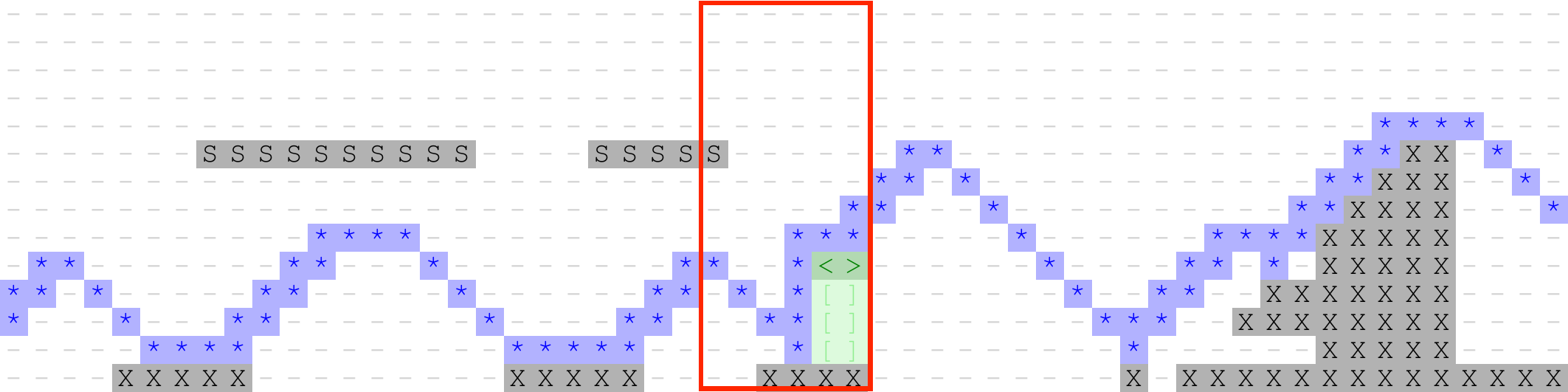}
      \\
    \end{tabular}
    \caption{ \label{XFigureMarioLinkers} Two \textit{Mario} segments being linked together. The red line represents where a linker could be placed. Purple stars represent a path that can be taken by the player to complete the level. Null linker does not result in a completable level.}
\end{figure}
}

\newcommand{\XFigureIcarusLinkers}{
\begin{figure}
    \centering
    \small
    \begin{tabular}{ccc}
       \includegraphics[width=0.8in]{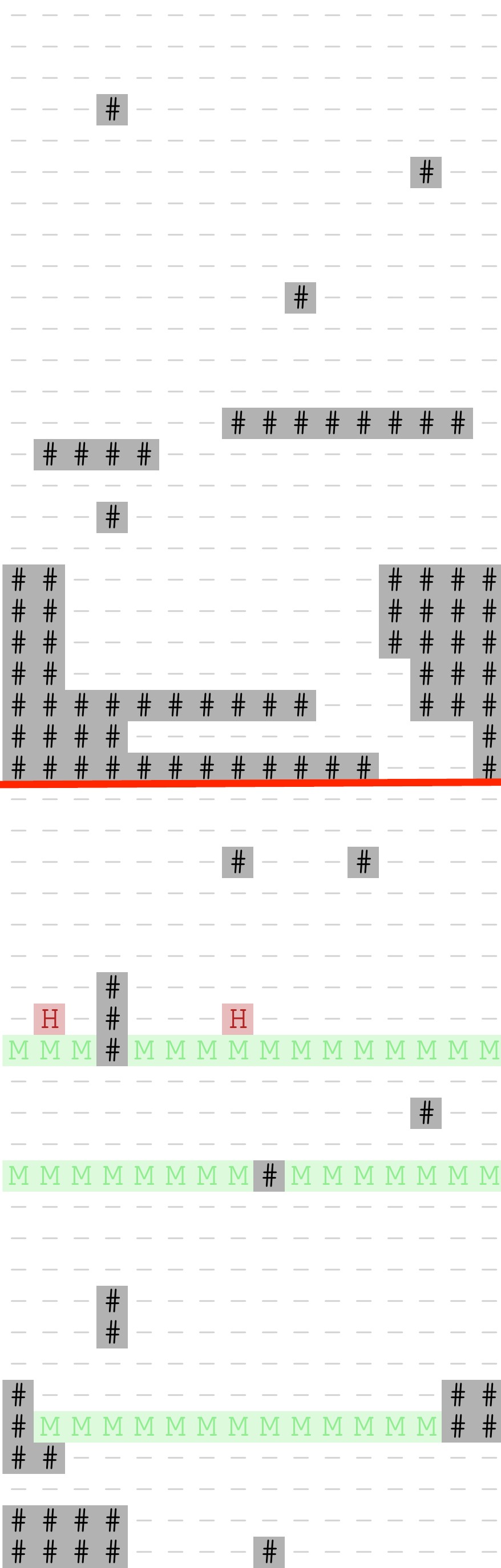}
       & 
       \includegraphics[width=0.8in]{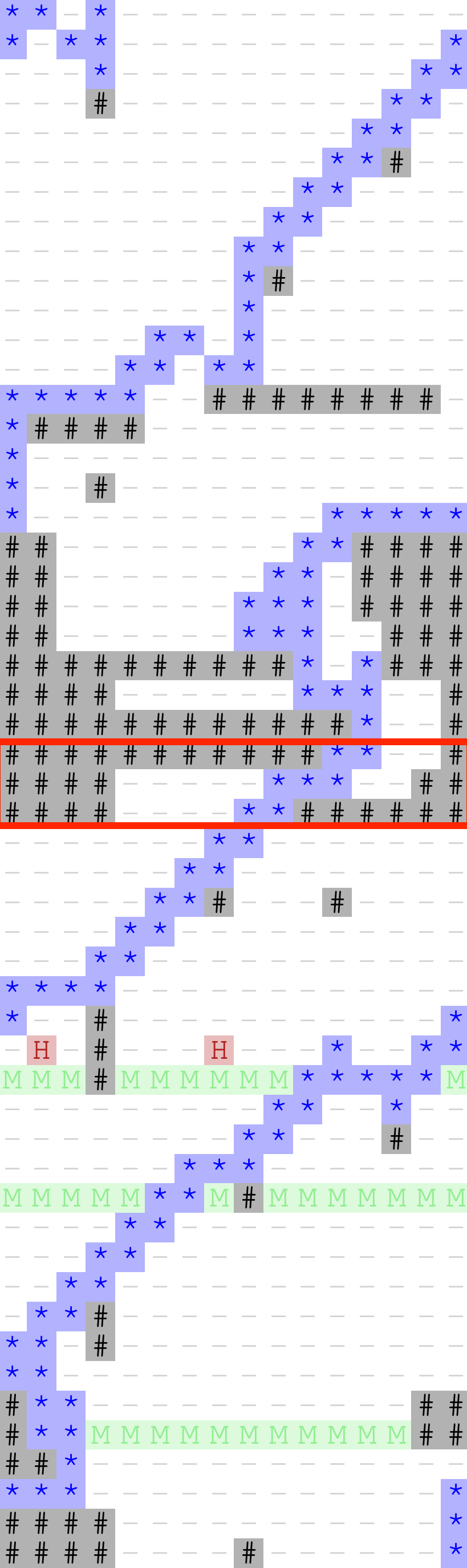}
       &
       \includegraphics[width=0.8in]{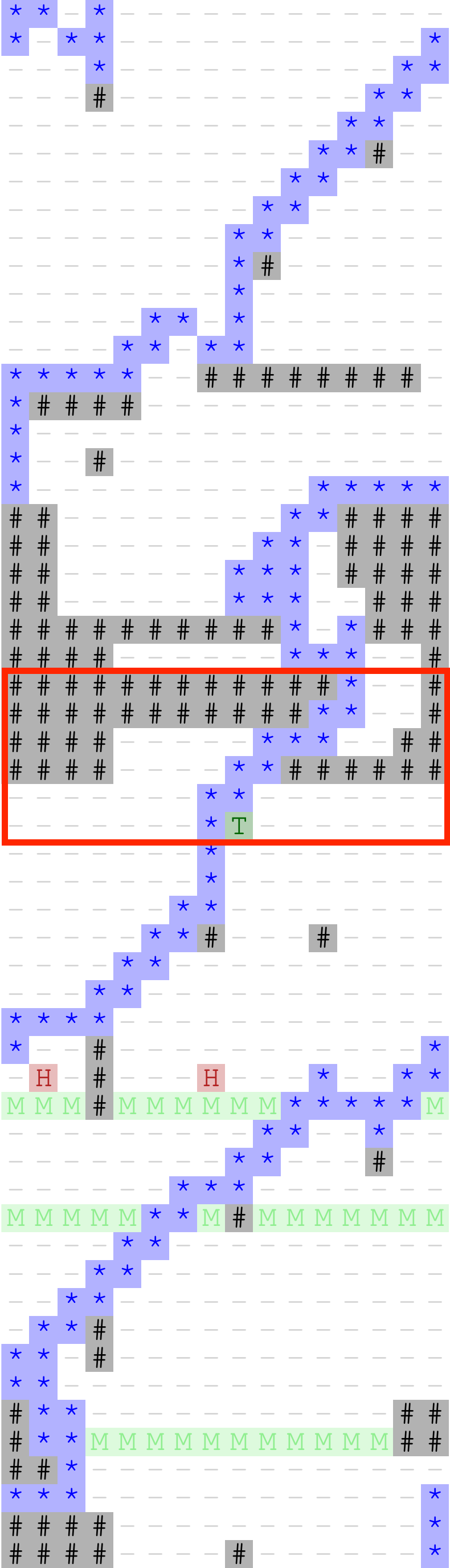}
       \\
       Null & Shortest & BC-match
    \end{tabular}
    \caption{ \label{XFigureIcarusLinkers} Two \textit{Icarus} segments being linked together. The red line represents where a linker could be placed and the red box represents a linker built. Purple stars represent a path that can be taken by the player to complete the level. Null linker does not result in a completable level. }
   
\end{figure}
}

\newcommand{\XFigureNullSuccess}{
\begin{figure}
\centering
\small
\begin{tabular}{cl}
    \rotatebox{90}{~~~~~Null} &   \includegraphics[height=0.6in]{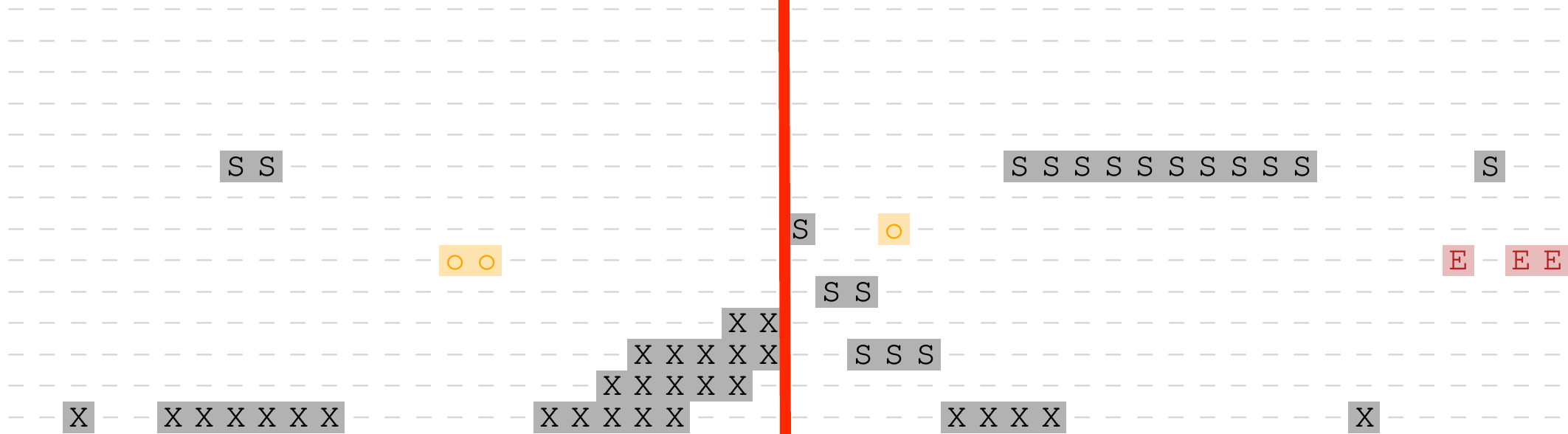} \\
    \rotatebox{90}{\shortstack{N-Gram\\Solution}} & \includegraphics[height=0.6in]{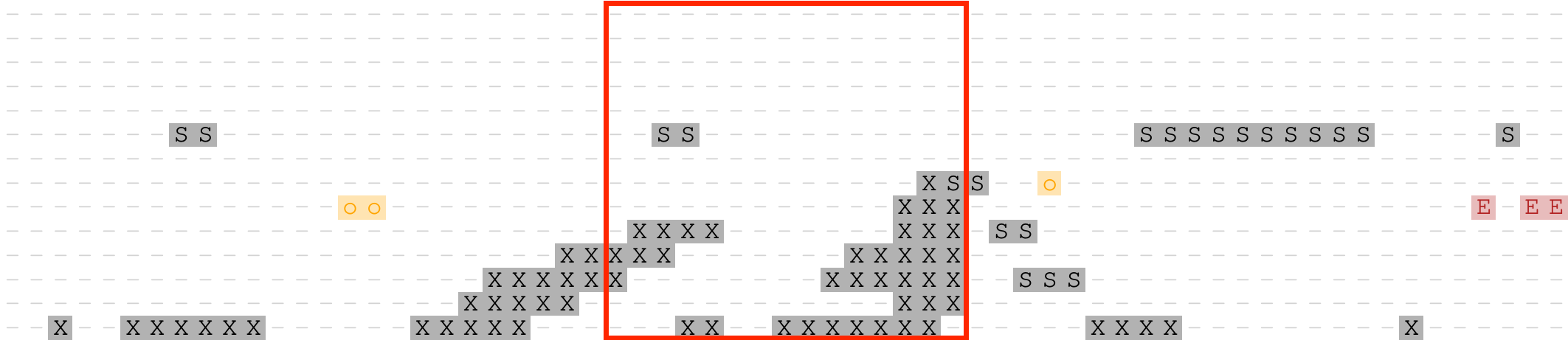}
\end{tabular}

\caption{\label{XFigureNullSuccess} Example of linking two \textit{Mario} segments where the depth-limited linkers cannot find a solution at depth six. (top) A null linker is completable but not generable. (bottom) At depth fifteen, a solution can be found.}
\end{figure}
}

\newcommand{\XFIGUREStructures}{
\begin{figure}
    \centering
    \begin{tabular}{ccc}
        \includegraphics[width=0.75in]{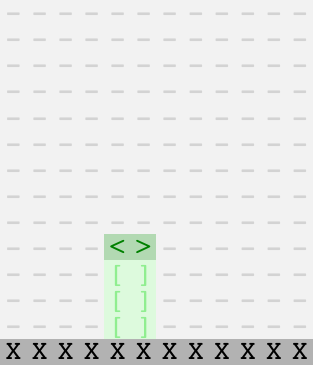} &
        \includegraphics[width=1.00in]{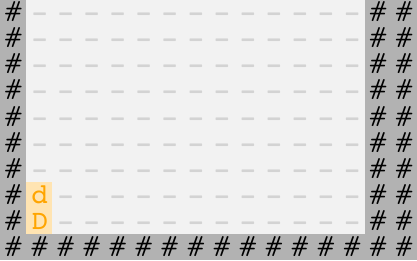} &
        \includegraphics[width=0.75in]{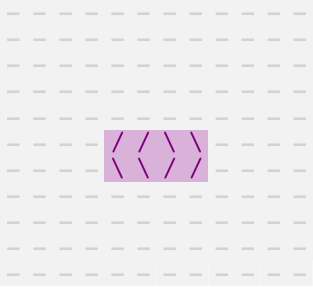} \\
        (a) \textit{Mario}  & (b) \textit{Icarus} & (c) \textit{DungeonGrams}
    \end{tabular}
    \caption{\label{XFIGUREStructures} Examples of structures for all three test games. \textit{Mario} has pipes which are two columns wide and some number of rows tall. \textit{Icarus} has doors that are one column wide and two rows tall. \textit{DungeonGrams} has a structure that is 4 columns wide and 2 rows tall.}
\end{figure}
}

\newcommand{\XFIGUREMarioLinkingExample}{
\begin{figure}
    \centering
    \setlength{\tabcolsep}{2pt}
    \begin{tabular}{rl}
        \rotatebox{90}{\strut~~~Concat.} &
        \includegraphics[height=0.75in]{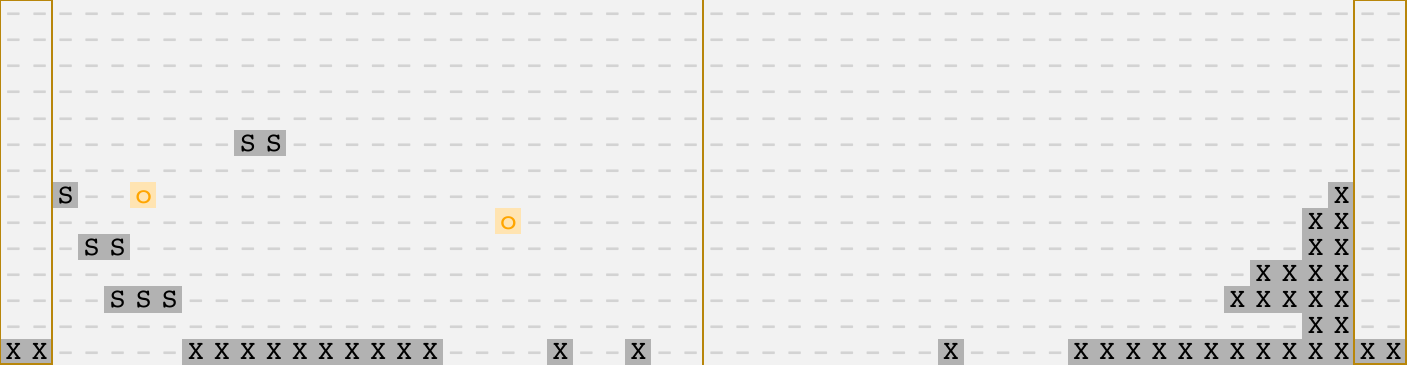} \\
        \rotatebox{90}{\strut~~~Linking} & 
        \includegraphics[height=0.75in]{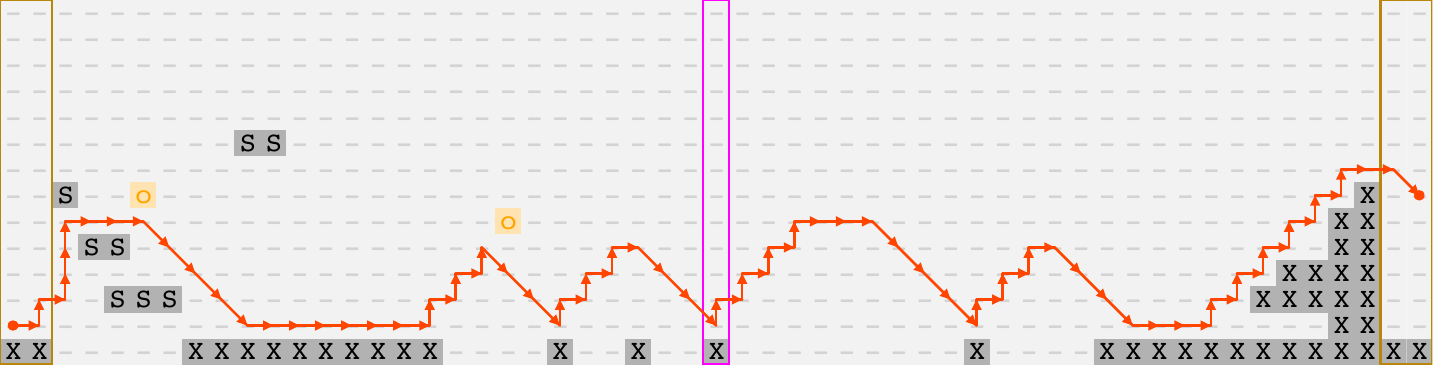} 
        \\[10pt]
        \rotatebox{90}{\strut~~~Concat.} &
        \includegraphics[height=0.75in]{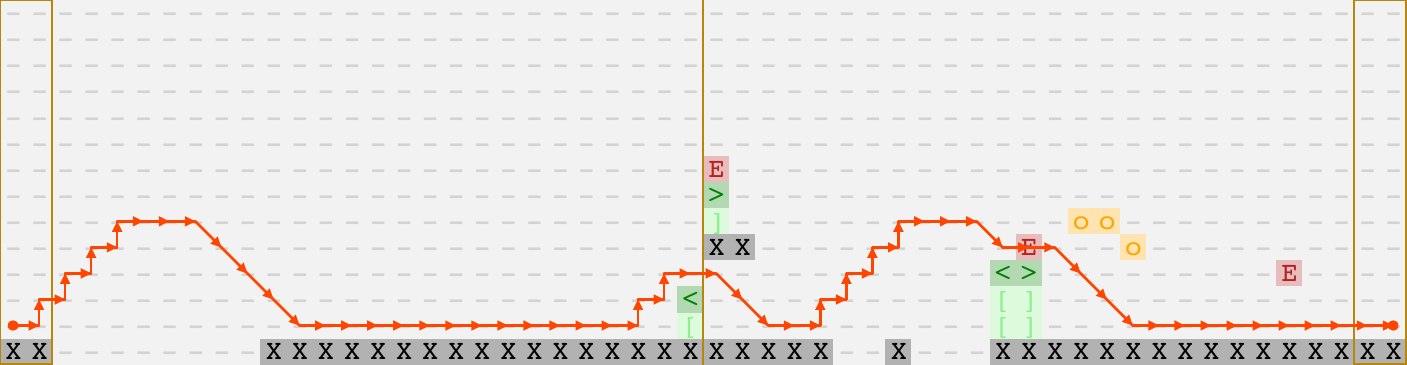} \\
        \rotatebox{90}{\strut~~~Linking} & 
        \includegraphics[height=0.75in]{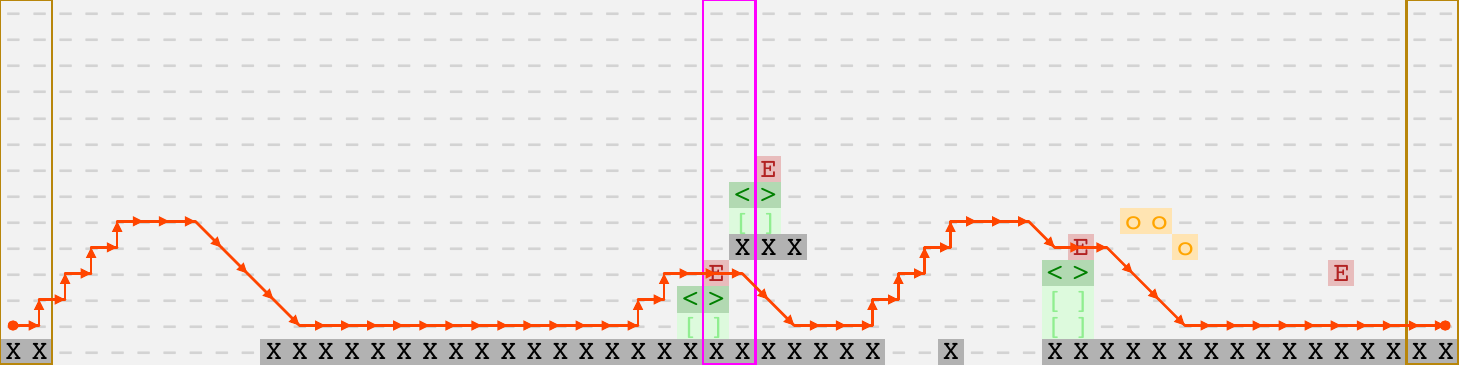} 
    \end{tabular}
    \caption{\label{XFIGUREMarioLinkingExample} Two examples for \textit{Mario}. The example on top shows concatenation failing due to a large gap. The bottom displays the longest linker produced. The tan line in the middle is where the link would have been placed for concatenation. The tan boxes on the left and right show the padding. The magenta box in the middle shows the linker found. The red lines show a path through the level if it was beatable.}
\end{figure}
}

\newcommand{\XFIGUREIcarusLinkingExample}{
\begin{figure}
    \centering
    \begin{tabular}{ccccc}
        \includegraphics[width=0.65in]{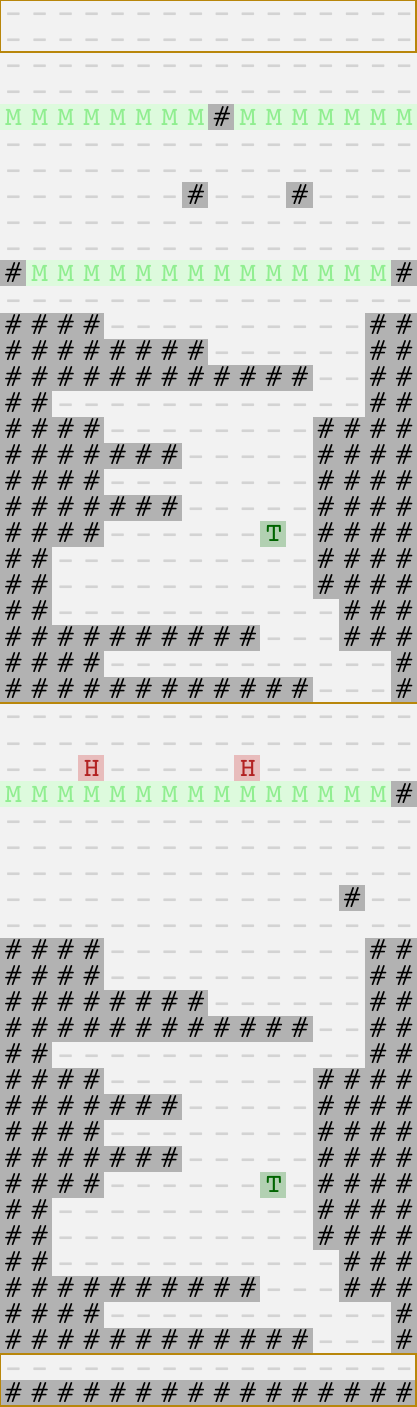} &
        \includegraphics[width=0.65in]{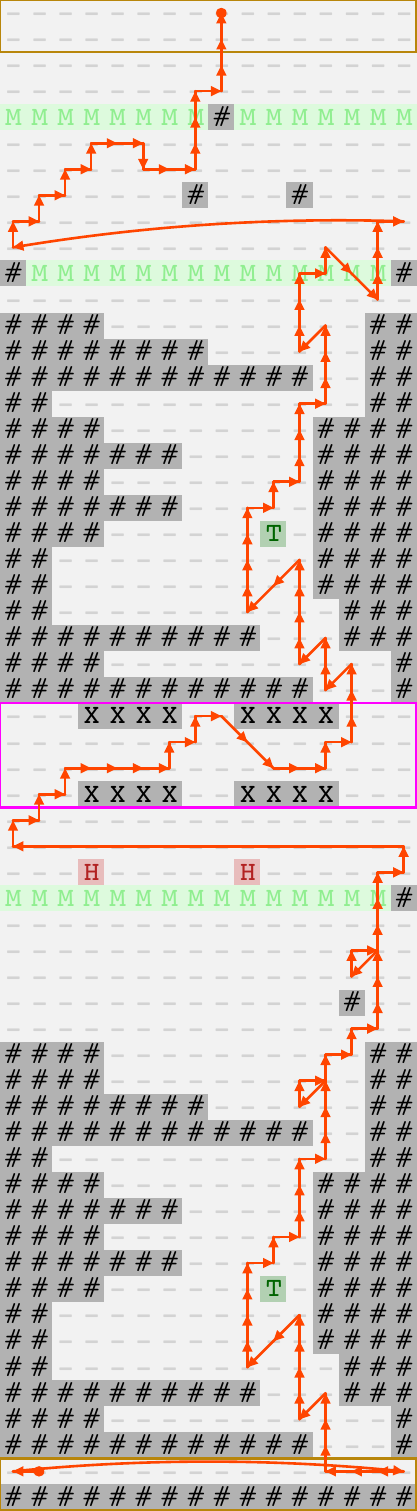} &&
        \includegraphics[width=0.65in]{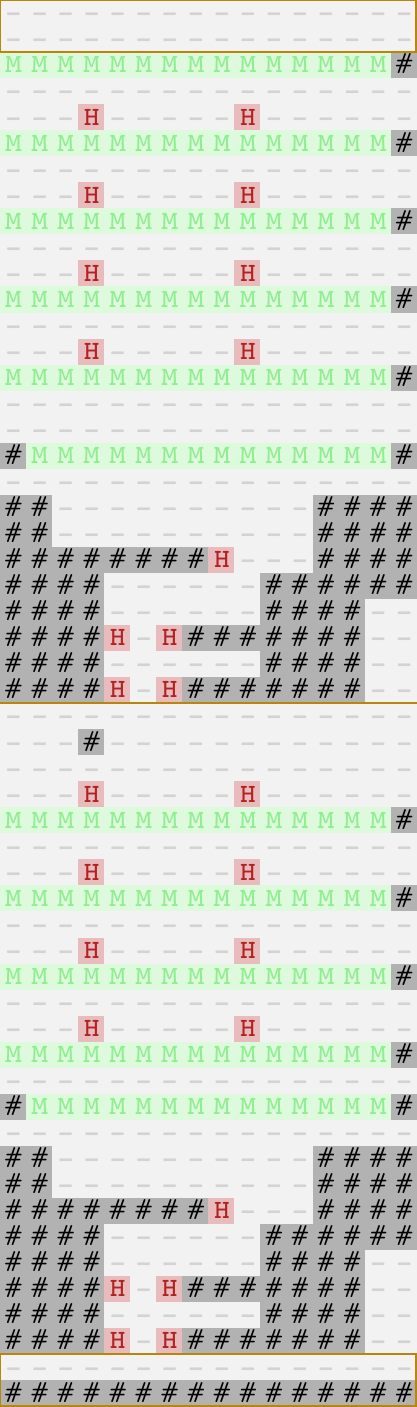} &
        \includegraphics[width=0.65in]{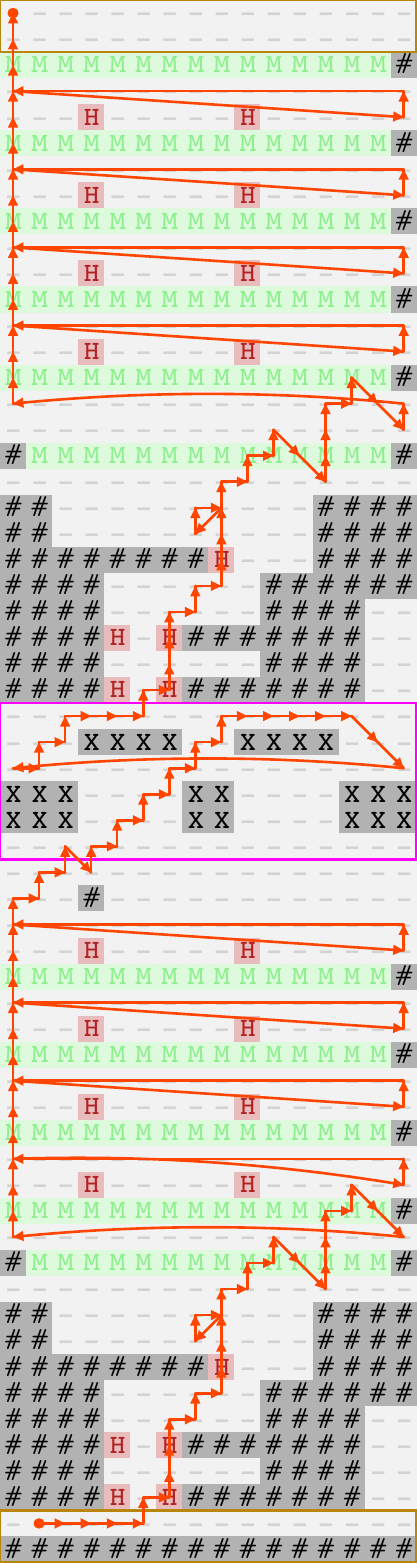} 
        \\
        Concat. & Linking && Concat. & Linking
    \end{tabular}
    \caption{\label{XFIGUREIcarusLinkingExample} Two examples for \textit{Icarus}. Both show concatenation failing due there being no possible path to complete the level. The right example of concatenation and linking shows the longest linker required to make a level completable. Padding is shown with the tan boxes at the top and bottom of the level. The tan line in the middle is where the link would have been placed for concatenation. The magenta box shows the linker found. The red lines show a path through the level if it was beatable.}
\end{figure}
}

\newcommand{\XFIGUREDGLinkingExample}{
\begin{figure}
    \centering
    \setlength{\tabcolsep}{2pt}
    \begin{tabular}{rl}
        \rotatebox{90}{\strut~~~Concat.} &
        \includegraphics[height=0.75in]{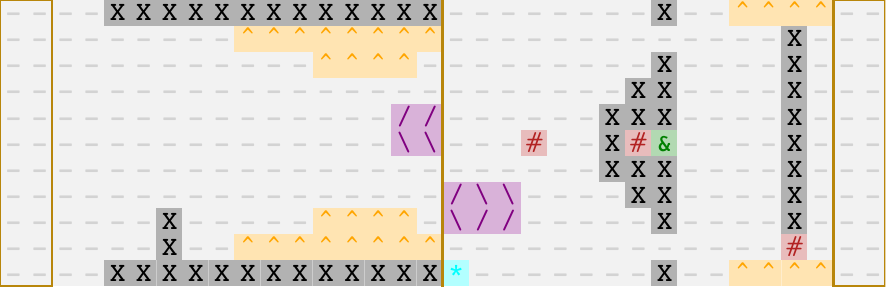} \\
        \rotatebox{90}{\strut~~~Linking} & 
        \includegraphics[height=0.75in]{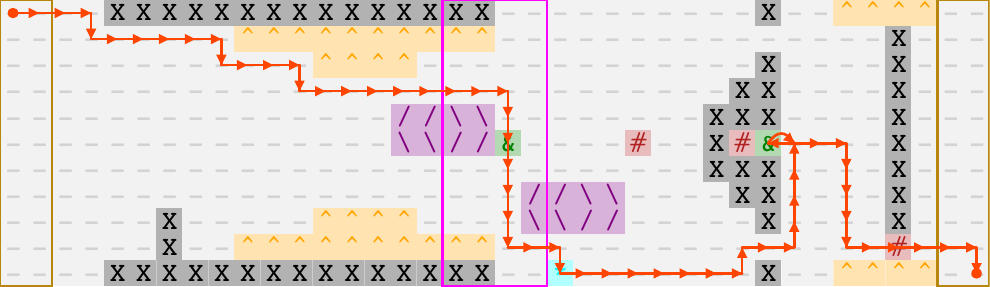} 
        \\[10pt]
        \rotatebox{90}{\strut~~~Concat.} &
        \includegraphics[height=0.75in]{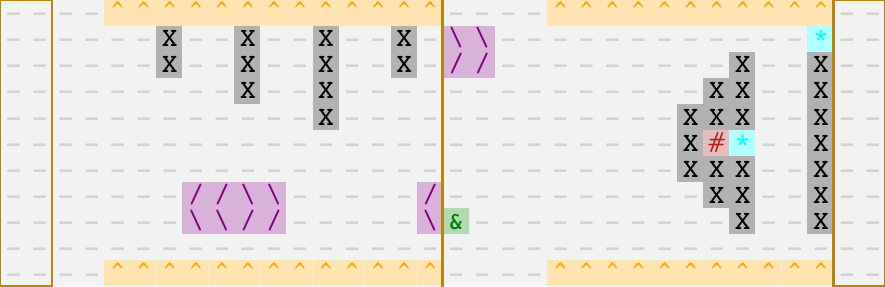} \\
        \rotatebox{90}{\strut~~~Linking} & 
        \includegraphics[height=0.75in]{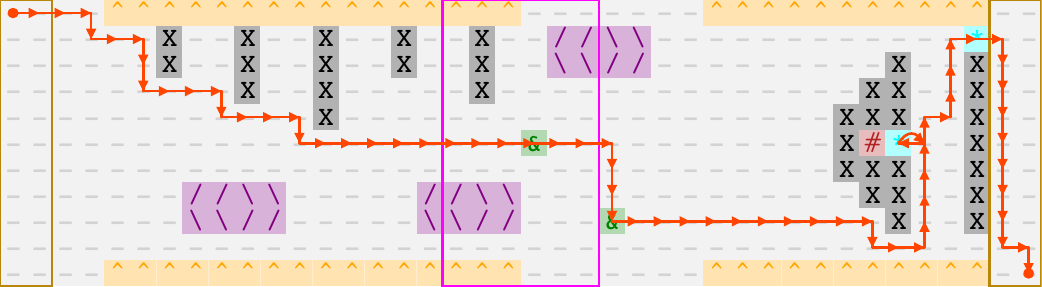} 
    \end{tabular}
    \caption{\label{XFIGUREDGLinkingExample} Two examples for \textit{DungeonGrams} where concatenation fails due to the agent running out of stamina. ($\&$ represents food.) The bottom concatenation and linking example shows the largest linker found. The tan boxes on the left and right show the padding. The tan line in the middle is where the link would have been placed for concatenation. The magenta box shows the linker found. The red lines show a path through the level if it was beatable}
\end{figure}
}

\newcommand{\XFIGUREDGUncompletable} {
\begin{figure}
    \centering
    \begin{tabular}{rl}
        \rotatebox{90}{~~~~DG} &
        \includegraphics[height=0.63in]{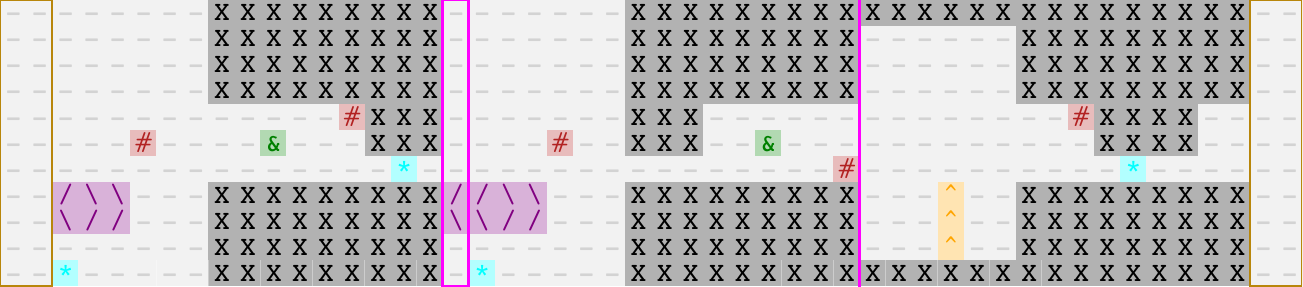} \\
        \rotatebox{90}{~DG-Food} & 
        \includegraphics[height=0.63in]{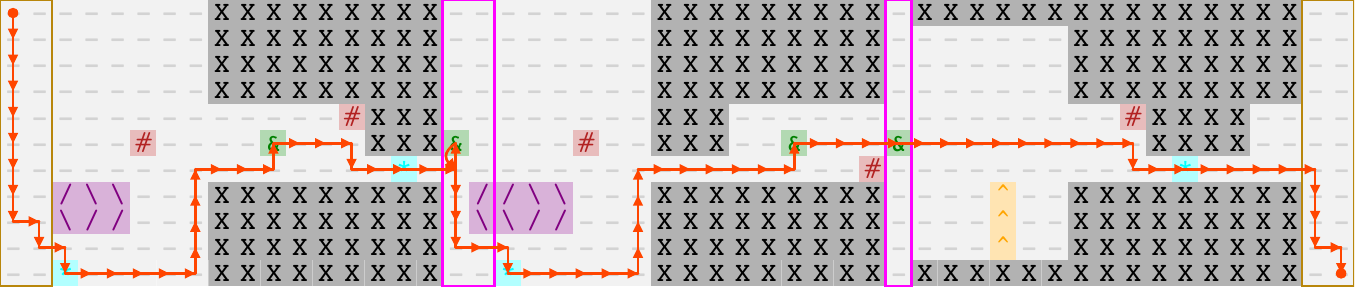} 
    \end{tabular}
    \caption{\label{XFIGUREDGUncompletable}Example of three segments being linked in \textit{DungeonGrams}. Tan boxes on either side represent padding. Magenta boxes in the middle represents a linker and the magenta line shows that no linker was necessary for the second and third level segment for linking (\textit{DG}). The agent fails to reach the end for \textit{DG} due to a lack of food, but this was not a problem for linking with food required (\textit{DG-Food}).}
    
\end{figure}
}


\newcommand{\z}{\hphantom{0}}
\newcommand{\x}[1]{\textbf{##1}}
\newcommand{\XTABLELinksFound}{
\begin{table}[t]
\centering
\small
\setlength\tabcolsep{2pt}
\renewcommand\arraystretch{2}
\begin{tabular}{|c|c|>{\centering}p{1.7cm}|>{\centering}p{1.7cm}|>{\centering\arraybackslash}p{1.7cm}|}
    \hline
     & Type & Unbroken & Completable & Usable \\
    \hline
    \hline
     \multirow{2}{*}{\rotatebox{90}{\textit{Mario}}} &
    Concatenation
      &  0.97 & 0.92 &0.89
     \\
    \cline{2-5}
    &
    Linking
      &  1.00 & 1.00 &1.00
      \\
    \hline
    \hline
    \multirow{2}{*}{\rotatebox{90}{\textit{Icarus}}} &
    Concatenation
      & 0.96 &0.11 & 0.11
     \\
    \cline{2-5}
    &
    Linking
      &  1.00 & 0.99 & 0.99
      \\
    \hline
    \hline
    \multirow{2}{*}{\rotatebox{90}{\textit{DG}}} &
    Concatenation
      & 0.52 & 0.58 & 0.23
     \\
    \cline{2-5}
    &
    Linking
      &  1.00 & 1.00 & 1.00
      \\
    \hline
\end{tabular}
\vspace{2pt}
\caption{\label{XTABLELinksFound} Percentage of links that resulted in unbroken levels, completeable levels, and usable levels. DG is short for \textit{DungeonGrams}.}
\end{table}
}

\newcommand{\XTABLECombining}{
\begin{table}[t]
\centering
\small
\renewcommand\arraystretch{1.2}
\setlength\tabcolsep{2pt}
\begin{tabular}{|c|c|c|c|c|}
\hline
 Segments & \textit{Mario} & \textit{Icarus} & \textit{DG} & \textit{DG-Food}
 \\
 \hline
 2 & 1.00 & 1.00 & 1.00 & 1.00
 \\
 \hline
  3 & 1.00 & 1.00 & 0.99 & 0.99
 \\
  \hline
  4 & 1.00 & 1.00 & 0.98 & 0.99
 \\
  \hline
  5 & 1.00 & 1.00 & 0.97 & 0.99
 \\
 \hline
\end{tabular}
\vspace{2pt}
\caption{\label{XTABLECombining}Shows the percentage that linked levels are usable. DG is short for \textit{DungeonGrams} where an empty linker is allowed. \textit{DG-Food} requires that the tree search adds at least one level slice to the linker.}
\end{table}
}

\newcommand{\XTableLinkStats}{
\begin{table}[]
    \centering
    \renewcommand\arraystretch{1.2}
    \setlength\tabcolsep{2pt}
    \begin{tabular}{|c|c|c|c|c|c|c|}
    \hline
        Game & \shortstack{\\Mean\\Length} & \shortstack{\\Median\\Length} & \shortstack{\\Max\\Length} & \shortstack{\\Mean\\$D_{BC}$} & \shortstack{\\Median\\$D_{BC}$} & \shortstack{\\Max\\$D_{BC}$} \\
        \hline
        \textit{Mario}  & 0.110 & 0 & 2 & 0.001 & 0.000 & 0.020 \\
        \textit{Icarus} & 2.658 & 3 & 6 & 0.008 & 0.008 & 0.029 \\
        \textit{DG}     & 1.147 & 1 & 6 & 0.012 & 0.011 & 0.061 \\
        \hline 
    \end{tabular}
    \vspace{2pt}
    \caption{For each game, shows the mean, median, and max for lengths of linkers found and $D_{BC}$. DG is short for \textit{DungeonGrams}.}
    \label{XTableLinkStats}
\end{table}
}

\newcommand{\XTableMarioCompletability}{
\begin{table}[t]
\centering
\small
\setlength\tabcolsep{2pt}
\begin{tabular}{|c|c|c|c|c|c|}
\hline
& Min & Mean & Median & Max & std \\
\hline
no link, $k=2$ & 0.528 & 0.97 &1.0 & 1.0 & 0.104 \\
BFS, $k=2$   &  1.0  & 1.0  & 1.0 & 1.0 & 0.0 \\
best, $k=2$   &  1.0  & 1.0 & 1.0  &1.0 &0.0 \\
\hline
no link, $k=3$ & 0.359 & 0.938 & 1.0 & 1.0 & 0.167 \\
BFS, $k=3$   &  1.0  & 1.0  & 1.0 & 1.0 & 0.0 \\
best, $k=3$   &  1.0  & 1.0 & 1.0  &1.0 &0.0 \\
\hline
no link, $k=4$ & 0.262 & 0.899 &  1.0  & 1.0 & 0.216\\
BFS, $k=4$   &  1.0  & 1.0  & 1.0 & 1.0 & 0.0 \\
best, $k=4$   &  1.0  & 1.0 & 1.0  &1.0 &0.0 \\
\hline
\end{tabular}
\caption{\label{XTABLEplaceholder} Given that a set of links could be made for $k$ Mario segments, this shows how far the Summerville A* agent could traverse the linked segments. }
\end{table}
}



    

\newcommand{\XLISTINGAlgorithm}{
\begin{figure} 
\begin{algorithmic}
\Procedure{build\_link}{$start$, $end$}
    \For{s in forward\_chain.get($start$)}
        \For{e in back\_chain.get($end$)}
            \State $m \gets \text{tree\_search}(start + s,\: e + end)$
            \If{$m \neq \text{None}$}
                \State \Return $s + m + e$
            \EndIf
        \EndFor
    \EndFor
    \State{\Return None}
\EndProcedure
\end{algorithmic}
\caption{A pseudocode description for building a link.}\label{XLISTINGAlgorithm}
\end{figure}
}

    
\section{Introduction}

 The field of procedural content generation (PCG) researches algorithms that auto-generate content. A major focus of the field is the creation of levels~\cite{shaker2016procedural}. Approaches vary from graph grammars~\cite{hauck2020automatic} to machine learning \cite{summerville2018procedural} to reinforcement learning \cite{khalifa2020pcgrl} to behavior trees \cite{sarkar2021procedural} to search \cite{togelius2011search} and more. Generators generate whole levels or smaller segments that are combined to form larger levels. Segment generation is a promising area because it has potential to give designers more fine-grained control. To better understand this claim, consider the task of building a \textit{Mario} level with many jumps at the start, then a few enemies, then no jumps, and, finally, many jumps and many enemies. If a generator has to build the described level, then it has a large challenge due to the size of the search space. It can be more manageable to break the designer's requirements into parts by building smaller segments and combining them together. Unfortunately, little work has studied how to combine level segments to build full levels. This is important because the field works with games more complex than \textit{Mario}, and, as we will show, current approaches are insufficient for more complex games, and we need a better method.

The simplest approach to link level segments is \emph{concatenation}, or simply placing one segment after another \cite{volz2018evolving}.  Concatenation has no notion of completability or in-game structures (e.g. doors in \textit{Kid Icarus}). Another approach is to require padding on either side of a generated segment \cite{green2020mario}; this implicitly handles the problem of broken in-game structures. Green et al. extend this by using an agent to validate that a level is playable before showing it to the user \cite{green2020mario}. However, these examples only use \textit{Mario}.

As an alternative to concatenation, we present the process of \emph{linking} for 2D tile-based games, which creates a \emph{linker}---a tiny level segment that connects two level segments. The full level is the concatenation of the first segment, the linker, and the second segment. The full level should be \emph{completable} (i.e. a player can get from the beginning to the end) and \emph{unbroken} (i.e. a level should contain no broken in-game structures). A linker is \emph{usable} if it results in a level that is completable and unbroken. 

To generate usable linkers,\footnote{\href{https://github.com/bi3mer/LinkingLevelSegments}{https://github.com/bi3mer/LinkingLevelSegments}} we present a method of linking which uses two Markov chains and a tree search. This method relies on breaking a set of input levels into vertical or horizontal level slices. These level slices are used as input for two Markov chains. Both chains are used for structure completion for their respective level segment to guarantee an unbroken level by adding level slices to the linker---a limitation is that Markov chains can fail for unseen input. If we concatenate with the linker generated by the Markov chains, we only have a guarantee that a generated linker will result in an unbroken level. To address completability. we use a tree search which runs on the input level slices or a set defined by the designer. The output of the search is an ordered list of level slices---the list may be empty. These are added in between any level slices added by the two Markov chains to form the linker. If a linker can be produced, it is guaranteed to be usable. 

We test  with three games: \textit{Mario}, a horizontal platformer; \textit{Kid Icarus}, a vertical platformer; and \textit{DungeonGrams}, a top-down roguelike.\footnote{\href{https://github.com/crowdgames/dungeongrams}{https://github.com/crowdgames/dungeongrams}} For all three games, we test linking with level segments previously generated by Gram-Elites in past work \cite{biemer2021gram}.\footnote{\href{https://github.com/bi3mer/GramElitesData}{https://github.com/bi3mer/GramElitesData}} We choose \textit{Mario} as a baseline because much research in procedural content generation via machine learning \cite{summerville2018procedural} for games uses it. \textit{Icarus} is a platformer but has more complicated vertical pathing than \textit{Mario}. \textit{DungeonGrams} is a different genre, which means it has distinct requirements and can show that our method works for more than just platformers.

We compare concatenation with our linking approach. Linkers always result in unbroken levels. We also find that linking is always able to find usable linkers between two segments for \textit{Mario} and \textit{DungeonGrams}. Concatenation works well for \textit{Mario}, but often fails for \textit{DungeonGrams}---structures are broken, the level is not completable, or both. For \textit{Kid Icarus}, our approach almost always finds usable linkers and concatenation is likely to fail. We extend testing linkers between only two segments by using them to form larger levels consisting of multiple linked segments. For \textit{DungeonGrams}, we find that increasing the number of segments can result in levels that are not usable. For \textit{Mario} and \textit{Icarus}, though, increasing the number of segments does not affect usability. 

\section{Related Work}

This section is broken into four parts. First, we look at previous work that has combined vertical slices and tree search to generate levels. Second, we consider past work which connects dungeon rooms to form larger dungeons. Third, we discuss the similarity of our work to level repair and the work done in the area. Lastly, we examine previous work that has combined level segments to form larger levels.

\subsection{Markov Chains and Tree Search}

2D tile-based game levels can be broken down from a grid to an ordered set of horizontal or vertical slices. Dahlskog et al. \cite{dahlskog2014linear} use these slices as input to an n-gram \cite{jurafsky2000speech} to generate levels. Summerville et al. expand on this by using slices as input for a Markov chain where the output can be more than one level slice \cite{summerville2015mcmcts}. They use Monte-Carlo tree search (MCTS) \cite{coulom2006efficient} to generate full levels. This addresses a weakness of generation by following probabilities in the Markov chain: there is no guarantee on what will be produced. The addition of tree search allows Summerville et al. to produce levels that are completeable and unbroken. Further, they guide the search to match target characteristics.

\subsection{Connecting Dungeon Rooms}

Liapis \cite{liapis2017multi} uses evolution to generate a higher-order representation of a dungeon based on rooms. A secondary evolutionary process is run to build these rooms where every room has constraints (e.g. must have a connection to the left and right). These constraints are how rooms, or level segments, are correctly linked to form a larger level.

An alternative approach comes from Ashlock and McGuinness \cite{ashlock2014automatic}. They use evolution to fill in a grid with different tile types. Dynamic programming is built into the fitness function to guarantee that a path exists between all checkpoints in the level, which ensures that the larger level will be completable. They then use a room membership algorithm to find the rooms that were evolved and a separate algorithm to find adjacent rooms. Lastly, they fill in the rooms with content based on required and optional content. 

\subsection{Level Repair}

One way to view this work is that we are proposing a repair agent for a level formed from level segments. Cooper and Sarkar propose one approach to level repair using a pathfinding agent that can find a path that is impossible following the game's mechanics \cite{cooper2020pathfinding}. Points in the path that are impossible are repaired by, for example, removing a solid block that the agent jumped through.

Another approach comes from Zhang et al. \cite{zhang_video_2020}, where they use a GAN \cite{goodfellow2014generative} to generate a level, but find that it does not reliable encode playability. To address this they use a ``generate-then-repair'' framework. Levels are generated with the GAN, and then repaired with mixed integer linear programming, which finds the minimum number of changes that result in a fully playable level. 

Jain et al. \cite{jain2016autoencoders} use an autoencoder trained on \textit{Mario} levels from the VGLC~\cite{summerville2016vglc}. Among other uses, the autoencoder network is used to repair levels. In their work, a level is broken into windows or small level segments which can be input into the network. By placing a broken window into the network, they show that the network will output a similar segment that is playable.

\subsection{Using Level Segments}

Volz et al. \cite{volz2018evolving} trained a generative adversarial network (GAN) \cite{goodfellow2014generative} on a single \textit{Mario} level. Their innovation in the process was to search the latent space of the generative network with CMA-ES \cite{hansen2003reducing}. In one of their experiments, they tested forming larger levels with increasing difficulty. This was accomplished by generating the level segments with a target difficulty and then concatenating them together, which means there is no guarantee that the larger level is unbroken or completable.

Sarkar and Cooper \cite{sarkar2020sequential} use variational autoencoders (VAE) \cite{kingma2013auto} trained on levels with path information to generate level segments. The VAE was trained to encode a level segment and then use the encoding to predict the segment that follows. A level is generated by starting with one segment and then sequentially building new segments with the VAE. Meaning, the VAE is performing the role of generating segments sequentially in a way that they connect to each other.

Green et al. \cite{green2020mario} start with a corpus of \textit{Mario} levels that have been generated with a required padding of two vertical slices of ground tiles on either side \cite{khalifa2019intentional}. The segments are linked together with the padding to form larger levels. Larger level sequences are found by using FI-2Pop \cite{kimbrough2005introducing}. Infeasible sequences are optimized to be completable. Feasible sequences are optimized to match a target mechanics sequence. The padding guarantees unbroken levels and FI-2Pop finds combinations of segments that are completable.

Li et al. \cite{li2021ensemble} use an ensemble of Markov chains to generate \textit{Mega Man} levels. They use a first-order Markov chain to
model the direction (i.e. horizontal or vertical) of a set of rooms. They use two L-shape Markov chains \cite{snodgrass2014hierarchical} to generate the rooms, one for horizontal rooms and the other for vertical rooms. They place these rooms together and run a check to guarantee that a path from one to the other exists. If a valid path does not exist, they re-sample until two rooms with a connection are found, guaranteeing a completable level.

\section{Approach}

\XLISTINGAlgorithm

Here, we describe our approach. First, we give a brief overview of Gram-Elites, which previously generated the level segments we use.  Second, we describe our approach to link two segments together. Finally, we review the games used in this work.

\subsection{Gram-Elites}

Gram-Elites \cite{biemer2021gram} is an extension to MAP-Elites \cite{mouret2015illuminating} that uses n-grams for population generation and the genetic operators mutation and crossover. These operators use a concept called \textit{connection}, which runs a breadth-first search through the n-gram to ensure that post-modification the new segment is generable by the n-gram. Like MAP-Elites, Gram-Elites uses behavioral characteristics \cite{smith2010analyzing} to differentiate level segments---we give a brief description of the behavioral characteristics used for each game in section \ref{ssec:games}. Gram-Elites guarantees that all output segments are generable by an n-gram and optimizes segment completability. 

For this work, we filter out segments that were not completable.

\subsection{Generating Linkers}

When building a linker between two segments, there are three problems. First, does the starting segment end with an incomplete structure? (Figure \ref{XfigureMalformedPipe}a is an example of this problem.) Second, does the end segment begin with an incomplete structure? Third, does the linker result in a completable level when used to link the two segments? We start by addressing the first two problems.

One option to address level segments that start or end with incomplete structures is to filter them out, but we believe a more flexible approach is called for. We address this with structure completion. In some games, like \textit{Mario}, it would be simple to check for incomplete pipes at the edge of segments and add a custom level slice to complete them. To be more general, we take advantage of using levels that are generable by an n-gram. We use two Markov chains to address incomplete structures. (Note that these chains do not modify the level segments themselves and do not act as level repairers.) The first, the forward chain, addresses potentially incomplete structures at the end of the starting segment. This chain is initialized from data in the direction the designer meant (e.g. \textit{Mario} is from left to right). The second, the back chain, reads in reverse. We filter the data each Markov chain receives to only include level slices with tiles associated with game structures, resulting in input and output that can be greater than one level slice. Further, input is limited by the size of a structure---e.g. \textit{DungeonGrams} structures are four columns long---to prevent the potential problem of adding additional structures where only one is called for.

\XfigureMalformedPipe

Figure \ref{XfigureMalformedPipe} shows the forward chain in action. It uses the last slice of the starting segment as input into the chain. If there is no output, there are no incomplete structures at the end of the starting segment. If there is output, check if there is output for the last two slices. If so, check for three, and so on until there is no output. We use the output associated with the most slices in the Markov chain. The back chain follows the same process, but in the opposite direction. Once both chains run, we have a guarantee that the concatenation of the starting segment, forward link, back link, and end segment has no incomplete structures if the input segments are generable by the n-gram used in Gram-Elites, which comes from the data used to create the forward and backward Markov chains. If either chain receives a structure not present in the input data, structure completion with Markov chains will fail. 

The next step is to address the problem of guaranteeing that a linked level is completable. \textit{Linking slices} are level slices which do not contain in game structures. By default, these are the level slices not input into the previously described Markov chains. Alternatively, the designer can define them. A breadth-first search uses linking slices to find the minimum number of slices required to make the fully linked level completable. The output can be zero slices, but we can add constraints (e.g. require at least $k$ slices). We use a max-depth check to prevent an infinite search. A benefit of using designer-defined linking slices is that the search space is reduced and the algorithm is much faster.

View figure \ref{XLISTINGAlgorithm} for pseudocode of the proposed method. The for loops for both chains allow for the possibility that there may be multiple ways to complete a structure.

\subsection{Games} \label{ssec:games}

\XFIGUREStructures
Here we review the three games used to evaluate linking. When evaluating completability, we pad the beginning and end of levels being evaluated with a few level slices. This allows us to easily define where the agent starts and where the agent must reach for the level to be classified as completable.

\noindent
\\
\textbf{Mario} is a horizontal platformer. As input for the linker structures, we use training levels from the VGLC \cite{summerville2016vglc}. The level segments previously built with Gram-Elites use the same input levels, and the behavioral characteristics are linearity and leniency, which are the axes of the MAP-Elites grid. For the linking completability tree search, default linking slices were used with no designer intervention. As a reminder, the set of default linking slices does not include level slices with tiles related to pipes. The max depth for the linking tree search is 7. For our agent, we used a modified Summerville A* agent \cite{summerville2016vglc} to calculate the furthest point in the level that can be reached. The only in-game structures are pipes, see Figure \ref{XFIGUREStructures}a. 

\XFIGUREIcarusLinkingColumns
\noindent
\\
\textbf{Kid Icarus} is a vertical platformer. As input for the linker structures, we use training levels from the VGLC \cite{summerville2016vglc}. The level segments previously built with Gram-Elites use the same levels. The axes of the grid are density and leniency. For the linking completability tree search, we use one empty row and two rows with platforms designed to connect to anything else, see figure \ref{XFIGUREIcarusLinkingColumns}. The max depth for the linking tree search is 7. For our agent, we use a modified Summerville A* agent \cite{summerville2016vglc} to calculate the furthest point in the level that can be reached. \textit{Kid Icarus} has one structure which is a door, see figure \ref{XFIGUREStructures}b.

\noindent
\\
\textbf{DungeonGrams} is a top-down roguelike developed to evaluate Gram-Elites. The player has to progress from left-to-right and hit switches to unlock a portal and then use it as an exit to beat a level. There are spikes and enemies that the player has to avoid while navigating a level. \textit{DungeonGrams} also has as stamina mechanic where the player has a limited number of moves before they lose. The player can increase their stamina by finding food in the level.

As input for the linker structures, we use the 44 training levels made for Gram-Elites. The axes of the grid are density and leniency. For the linking completability tree search, we use three columns with food: bottom, middle, and top of each respective column. To evaluate completability, we use a search-based agent already built for \textit{DungeonGrams}. The max depth of the linking tree search is 4, which is lower than the other two games due to the agent being slower as a result of a larger search space. \textit{DungeonGrams} also has one structure that can be seen in figure \ref{XFIGUREStructures}c.

\section{Evaluation}

We break our evaluation into two parts. First we evaluate the effectiveness of concatenation and compare it to our linking algorithm. Second, we examine how well linkers made for two segments extend to linking multiple segments.

\subsection{Concatenation Versus Linking}

\XTABLELinksFound

Segments generated by Gram-Elites are organized by the MAP-Elites grid where each axis represents a behavioral characteristic \cite{smith2010analyzing}. Rather than selecting random segments and linking them together, we use the grid to select similar segments by using neighboring segments. There are $13,182$ possible links for \textit{Mario}, $9,453$ for \textit{Kid Icarus}, and $6,086$ for \textit{DungeonGrams}.

\XFIGUREMarioLinkingExample

Table \ref{XTABLELinksFound} shows the percentage of concatenations and links found that are unbroken, completable, and usable (i.e. both unbroken and completable). With \textit{Mario}, we can see that concatenation is $0.92$ likely to result in a level that is completable by an agent and $0.89$ likely to result in a level that is usable, showing that concatenation is likely to be an effective strategy for most use cases in a simple platformer. Linking is successful for every possible link. Figure \ref{XFIGUREMarioLinkingExample} shows two examples where concatenation fails. The top example shows that concatenation can fail when a gap is too large. The bottom example shows the largest linker required for \textit{Mario}. The linker completed two structures, but did not affect completability as the concatenated version was also completable. 

\XFIGUREIcarusLinkingExample

Concatenation performs well for \textit{Icarus} in terms of producing an unbroken level. However, it performs the worst at generating a completable level at just eleven percent. Linking, though, finds a usable link for almost every single starting and ending segment given. Figure \ref{XFIGUREIcarusLinkingExample} shows two examples where concatenation cannot produce a completable level. In both cases the agent does not have enough space to make the required jump. Tree search resolves this by adding two platforms for the agent. The right-most example shows the largest linker built.

\XFIGUREDGLinkingExample

Concatenation performs poorly for \textit{DungeonGrams}, and is unlikely to produce a usable level. This is due to two reasons. First, the size and complexity of the structures makes it unlikely for two segments to line up if either or both level segments have unfinished structures. Second, if the two segments do not have enough food, the concatenated level is impossible to beat. Linking, though, handles both problems and always produces a usable link. Figure \ref{XFIGUREDGLinkingExample} shows two examples where concatenation fails. In both cases, structures are incomplete and there isn't enough food for the agent. Linking succeeds by completing the structures with the forward and back chains, and tree search adds food in the middle to make the level completable. The bottom example shows the largest linker found for \textit{DungeonGrams}.

\XTableLinkStats

The usability of a linker isn't the only important metric to consider. Ideally, a linker should not be identifiable or distracting to the player (e.g. the same set of level slices between every segment). To examine this, we report on the lengths of linkers found and the change in behavioral characteristics of a two segment level when a linker is used. We determine the latter by finding the behavioral characteristics (BCs) of the concatenated level and the linked level, and calculate the euclidean distance between the two. The result is $D_{BC}$.

The results can be seen in Table \ref{XTableLinkStats}. For \textit{Mario}, we can see that a linker tends not to be required and, when it is, the worst case is a linker with two level slices. As a result, $D_{BC}$ must be small. To put the value in context, the implementation of Gram-Elites tessellates the grid by changes in BC every $0.025$ for \textit{Mario}. For \textit{Mario}, if a linked level and its corresponding concatenated level have a $D_{BC}$ smaller than $0.025$, then we consider them to be very similar. \textit{Icarus} is different in that it almost always requires a linker with two to three linking slices. The max length is also much longer than \textit{Mario}. The increment for \textit{Icarus's} grid is $0.0125$. In this case, the mean and median of $D_{BC}$ show that most linkers are unlikely to result in large changes to the BCs. The max $D_{BC}$ shows that the worst case is a jump of two to three neighbors in the grid. Finally, \textit{DungeonGrams} on average has smaller linkers than \textit{Icarus} but has the same max linker length. The mean and median of $D_{BC}$ is quite small: the Gram-elites grid is 0.05 per bin in \textit{DungeonGrams}.

Overall, linking usually results in minimal modifications to the player's experience per segment while providing the benefit of completability and unbrokenness.

\subsection{Linking Multiple Segments}

In this section, we test the linkers made for two segments by combining multiple segments with them. To test, for various numbers of segments, we generated one thousand random levels by following the MAP-Elites grid, neighbor by neighbor. Linkers were selected only if they had a playability guarantee for the starting and ending level segment. The results are in Table \ref{XTABLECombining}. Notably, we see that all two-segment levels are usable, as should be the case using the linkers.

For both platformers, we can see that increasing the number of segments does not change the results: all the linked multi-segment levels were usable.

\XTABLECombining
\XFIGUREDGUncompletable

For \textit{DungeonGrams}, we updated linking for a new approach. \textit{DG} refers to when an empty link is allowed. \textit{DG-Food} forces the link to have at least one linking slice from the tree search, meaning there is at least one column with food in the link since we use custom level slices which always have food. We find that \textit{DG} performs increasingly poorly when compared to the platformers as the number of segments increases. In contrast, the simple change for \textit{DG-Food} results in larger levels almost always being fully playable. Figure \ref{XFIGUREDGUncompletable} shows an example where \textit{DG} fails and \textit{DG-Food} succeeds. Failure occurs since linking guarantees that a path exists between two segments, but does not ensure that the player will have the same stamina at the start of a new segment as they do when starting the game. While food in \textit{DungeonGrams} does not give the player back their max stamina, it does increase the likelihood that the agent can make it through the next segment.

\section{Conclusion}

In this work, we present an alternative to concatenation for linking level segments of 2D tile-based games. Our approach uses two Markov chains and a tree search to guarantee that any linker produced results in a level that has no unbroken structures and is completable by an agent. We test on three games: \textit{Mario}, \textit{Kid Icarus}, and \textit{DungeonGrams}. 

When linking two segments, we find that our approach outperforms concatenation for all three games. In the two-segment case, it always finds a usable linker for both \textit{Mario} and \textit{DungeonGrams}. For \textit{Kid Icarus}, linking works for the vast majority of segments but can fail. On examining the failure cases, we believe this is partly due to the agent not perfectly playing the game by not correctly modelling blocks where the player can jump through and land on the top. We intend to improve the agent as part of future work.

We extend our evaluation and test how pre-built two-segment linkers work with sequences of level segments greater than two. We find that for \textit{Mario} and \textit{Kid Icarus}, all multi-segment levels are usable. For \textit{DungeonGrams}, though, performance is worse due to the stamina mechanic. We address this by requiring the link to always contain at least one column with food, and find that almost every large level can be completed. Another approach is to extend the linking strategy to link all $k$ segments simultaneously, where we currently only solve for $k=2$.

As it stands, our approach to linking can work with any segment-based generator if it generates segments that are generable by an n-gram; in this work, linkers used the same datasets that were used to generate the segments. As future work, we believe it is important to make a linking method that works for all generators, which means addressing the limitation of requiring n-gram generability. At the minimum, this requires updating structure generation to correctly handle unseen in-game structures. To further extend this work, we want to use the directed graph we built to evaluate linking, and use it to build larger levels tailored to individual players. 



\bibliographystyle{IEEEtran}
\bibliography{refs-manual}

\end{document}